\pdfoutput=1

\documentclass[11pt]{article}

\usepackage{ACL2023}
\usepackage{times}
\usepackage{latexsym}
\usepackage{graphicx}
\usepackage[T1]{fontenc}
\usepackage{placeins}

\usepackage[utf8]{inputenc}

\usepackage{microtype}

\usepackage{inconsolata}

\usepackage{array}

\title{Can Language Models Learn Analogical Reasoning? Investigating Training Objectives and Comparisons to Human Performance}

\author{Molly R. Petersen\textsuperscript{1,2} \and Lonneke van der Plas\textsuperscript{1,3} \\
  \textsuperscript{1} Computation, Cognition \& Language Group, Idiap Research Institute, Martigny, Switzerland \\
  \textsuperscript{2} NLP Lab, EPFL,  Lausanne, Switzerland \\
  \textsuperscript{3} Institute of Linguistics and Language Technology, University of Malta, Malta\\
  \texttt{molly.petersen@idiap.ch}, \texttt{lonneke.vanderplas@idiap.ch}\\}

\begin{document}
\raggedbottom
\maketitle

\begin{abstract}
While analogies are a common way to evaluate word embeddings in NLP, it is also of interest to investigate whether or not analogical reasoning is a task in itself that can be learned. In this paper, we test several ways to learn basic analogical reasoning, specifically focusing on analogies that are more typical of what is used to evaluate analogical reasoning in humans than those in commonly used NLP benchmarks. Our experiments find that models are able to learn analogical reasoning, even with a small amount of data. We additionally compare our models to a dataset with a human baseline, and find that after training, models approach human performance.

\end{abstract}

\maketitle

\section{Introduction}

Solving proportional analogies \(a:b::c:d\) (ex. \emph{Paris:France::Tokyo:Japan}) \cite{mikolov2013efficient,Mikolov2013DistributedRO,rogers-etal-2017-many} with embeddings has become an iconic way to demonstrate that they encode semantic information regarding the relationships between word pairs. Proportional analogies have been formalized with word embeddings by means of a simple arithmetic equation (often referred to as the vector offset method) \(b-a+c = d\) where the choice of \emph{d} is solved by maximizing \(cos(b-a+c,d)\), or other slightly more complicated similarity measures \cite{mikolov2013efficient,rogers-etal-2017-many,levy-goldberg-2014-linguistic,ushio2022bert}.

This task has developed into many datasets such as the Google Analogy Test Set and the Bigger Analogy Test Set (BATS) \cite{mikolov2013efficient,gladkova-etal-2016-analogy}. Examples of relation types included in these datasets are morphological in nature \((entertain : entertainer ::
examine : examiner)\), based on encyclopedic knowledge \((Paris:France:: Lima:Peru)\), or lexicographic relations, such as hypernyms and hyponyms \((candy:lollipop::color:white)\). Notably, the relations between entities that form each analogy are explicitly verbalized for each example, and are grouped into collections of many pairs with equivalent relations.
 While the ability of word embeddings to solve these sorts of analogies is interesting, the analogies contained in these datasets are different from what is typically used to test analogical reasoning in humans. Many would be trivially easy to solve, as is the case with morphological relations and a lot of encyclopedic knowledge \cite{ushio2022bert}. Even the task of completing an analogy by filling in a correct \emph{d} term is perhaps not well suited to humans. \citet{rogers-etal-2017-many} noted that analogy questions are often given to people as multiple choice questions, as people would likely fill in \emph{d} with a variety of words where a single correct answer is not always the case. It has also been pointed out that model performance on analogy tests tend to rely on how semantically related all the entities in the analogies are to each other, suggesting these tests might in fact not be measuring analogical reasoning \cite{rogers-etal-2017-many}.  When using the vector offset method, the \emph{a}, \emph{b}, and \emph{c} terms are generally excluded from candidates for \emph{d}. If they are not excluded \emph{b} or \emph{c} is often the closest neighbor to the estimate of \emph{d} and will be selected \cite{rogers-etal-2017-many,linzen-2016-issues}.

Analogical reasoning has the potential to extend beyond just solving proportional analogies. Analogies and analogical reasoning have value regarding scientific discovery, problem solving, human creative endeavors such as metaphor generation, as well as in the field of education \cite{boger2019,gentner2002analogy,gentner2016rapid,clement1988observed,gick1980analogical,gentner1997structure}.  
In this work, we do not focus on the semantic/morphological analogies in datasets such as BATS, but on more "complex" analogies that are closer to what is used to test analogical reasoning in humans, what \citet{ushio2022bert} referred to as "Psychometric Analogy Tests". Most of the data we use has been developed precisely to test humans. Furthermore, many of these analogies will have no verbalized relation between entities available, and there are likely no ways to group instances by type of relation without over-generalizing the relation. Unlike most previous studies, we want to go beyond just exploring what pretrained language models know about analogical reasoning. Our main contributions are as follows: we propose a training objective that allows language models to learn analogies. We find that learning is effective even with a fairly small amount of data, and approaches human performance on an unseen test set constructed for testing human analogical reasoning. Lastly, we find that fine-tuning models with this training objective does not deteriorate performance on external, but related tasks. All datasets are in English.

\section{Related Work}

While a lot of work tends to focus on performance on datasets such as BATS, there exist some datasets which were developed to test humans. 
\citet{ushio2022bert} explored the ability of word embeddings from various transformer models as well as some pretrained word embeddings (such as GloVe) to solve a variety of analogies originally designed to test humans in addition to the BATS and Google analogy test set. They tested three different scoring methods that aggregated across all valid analogy permutations. Their experiments found that the analogies designed for humans were much harder for all word embeddings to solve, with the better performing models' accuracy being less than 60\% on datasets designed for humans, while the highest accuracy attained on Google and BATS was 96.6\% and 81.2\% respectively.

\citet{czinczoll2022scientific} introduced a dataset composed of scientific and metaphorical analogies, called SCAN, also testing performance on transformer based embeddings by using a natural language prompt, and filling in an ending [MASK] token. They experimented with a zero and one-shot learning setting, and additionally fine-tuned models on the BATS dataset. Overall they found that performance on the SCAN was low, and that models which had been fine-tuned on BATS lead to a decreased performance on the SCAN dataset, which the authors attribute to the types of analogies being inherently different and that datasets such as the BATS did not well represent how humans utilize analogies.

In this work, we test a novel training objective to explore whether analogical reasoning in itself is a task that can be learned, and later test to unseen analogies with non-equivalent relation types. There have been a few works in addition to \citet{czinczoll2022scientific} that attempted at actively learning analogies or relations between words \cite{Ushio_2021, drozd2016word} using language models. \citet{czinczoll2022scientific} and \citet{drozd2016word}, handle the analogy problem in the more common way of predicting a word to complete the analogy. Our training focuses more on similarity in relations between entities, as opposed to similarity between entities in the analogy themselves, and uses this objective to fine-tune a pretrained BERT-base model to solve analogies as a classification problem. \citet{Ushio_2021} proposed RelBERT, whose motivation and training scheme is most similar to our set up in that they are focused on relations and relational similarity, even if their training task is not formulated as an analogy problem specifically. However their methods involve prompting and their training data was much larger (SemEval 2012 Task 2) and involved only lexical relations. Our work involves an arguably simpler training scheme, with a much smaller training dataset that includes a wider variety of analogical relations. 

 Additionally, there has been a lot of work in knowledge graph representation to learn embeddings for relations between entities, some of these which explicitly utilize analogical structure to learn embeddings \cite{liu2017analogical}. However, this area of research is out of scope as we are focusing specifically on learning to represent knowledge and relations in contextual language models.

\section{Methods} 
All our experiments used a pretrained BERT-base uncased model (110M parameters)\footnote{https://huggingface.co/} \cite{devlin-etal-2019-bert} unless stated otherwise. Models are trained on a binary classification task- when given an analogy, they must label it as either positive or negative.

We evaluate our models in three ways: 1) We evaluate the models' ability to learn analogy classification using two variants of a classification task, the binary classification task used during training and a ranking task, where we present them with both a positive analogy and its negative counterpart (how these are created is described in the \emph{Datasets} section below), much like a multiple choice question with two possible answers. Whichever pair the model scores as more likely an analogy is the one the model "chooses".  We chose to include this task as this is how it was formulated for the human baseline we use to compare our models, and is closer to how analogy tasks are often presented to humans \cite{rogers-etal-2017-many,jones2022differential}. Unlike the binary classification task, which will rely on a cutoff chosen to differentiate positive from negative, the decision made using ranking is relative to what a specific analogy is being compared to. 2) We compare the models' ability to solve analogies to a human baseline on an unseen dataset (also using ranking as described in the previous paragraph). 3) We evaluate performance on some external but related semantic similarity tasks. Statistical significance was determined using a two-sided z-test for proportions\footnote{Code available at https://github.com/idiap/analogy-learning}.

\subsection{SBERT-modifications}
Many of our experiments are a modification of the SBERT\footnote{https://www.sbert.net/} architecture presented by \citet{reimers2019sentencebert}. SBERT creates sentence level embeddings, using the sentence representations- as opposed to token level representations- to solve a variety of tasks such as sentence similarity. Sentence-level representations are created by feeding the full sentence through a BERT or BERT-variant model, then pooling the embeddings from the final hidden layer.

In the original SBERT setup, sentences are fed through a model and the token embeddings are pooled to create a single representation. The main modification we do is that instead of feeding SBERT sentences to create a single sentence-level embedding, we feed SBERT word pairs and create two word-level embeddings- one for each word in the pair. In case a word is separated into word pieces, this still requires a pooling layer. We mean-pool over the token level embeddings to create each word embedding.

Additionally, in the original paper, sentences were fed through individually. So, for example, in a sentence similarity task which rates the similarity between two individual sentences (no attending between the sentences), the representations are determined separately. Their two individual representations are then compared. In our specific case with the \(a:b::c:d\) analogy structure, the meaning of \emph{a} is dependent on \emph{b} and vice versa. However the meaning of \emph{c} is not necessarily related to \emph{a}, as they are just related to each other by their relation to their partner in their pair. Therefore we send through "\emph{a [SEP] b}" separately from "\emph{c [SEP] d}". This is only the case when the SBERT architecture is used. In the case that it is not, it will be explicitly stated.

\subsection{Models}
All models are trained\footnote{We did not do hyperparameter tuning as our goal is not to find the best model or beat any benchmarks, our goal is to test whether or not analogical reasoning is something that can be explicitly learned by comparing across models.} for three epochs, with a batch size of 32, and were trained using the Adam optimizer with an initial learning rate of $2e^-5$ \cite{loshchilov2019decoupled}. Each model took between ~2.5-5 hours to train on a single CPU. We use BERT-base in our main experiments, however reproduce several experiments with other BERT architectures in the appendix.

\subsubsection{Model 1: Simple Classifier}

As a first model, we trained BERT (not SBERT) to classify a proportional analogy as being an analogy or not. The model is fed the analogy in the form '\emph{a [SEP] b [SEP] c [SEP] d}', and outputs a 0 for not analogy or a 1 for true analogy. We chose to include one experiment with a simple classification head as this is the generic classification training paradigm for transformer models.

\subsubsection{Model 2: BERT a-b}

The rest of our models maintained formulating the task as a binary classification task, however instead of having a final classification head, we incorporated a novel training objective. We base our training off the idea that in a proportional analogy, the relationship between \emph{a} and \emph{b} can be framed as \emph{a-b} \cite{mikolov2013efficient}. Given that the pairs that make up an analogy should have the same relation expressed between the entities, it means \emph{(a-b)=(c-d)}.
We train our model to maximize the cosine similarity \emph{cos((a-b),(c-d))} for positive examples and minimize for negative. We use a cosine embedding loss to train the model, with a margin of 0.

\subsubsection{Model 3: BERT a-c}

Since the above experiments using cosine distance as a similarity measure do not take into account \(sim((a-c),(b-d))\), we test the model using \(cos((a-c),(b-d))\) as the training objective. This reordering is of interest because in these analogy pairs, \emph{a} and \emph{b} are semantically related to each other in some way (ex: \emph{nucleus} and \emph{electron}), but \emph{a} and \emph{c} are generally not (ex: \emph{nucleus} and \emph{sun}). This means that, in theory, \emph{a-c} is a much larger distance in semantic space. Previous studies have shown that when it comes to analogies, word embeddings often fail to capture more long-distance relations \cite{rogers-etal-2017-many}. Long-distance relations (connecting two seemingly unrelated concepts) is of interest to the topic of computational creativity.

\begin{table*}[ht]
\setlength\tabcolsep{2pt}
\begin{centering}

\begin{tabular*}{\textwidth}{@{\extracolsep{\fill}}c||*{3}{>{\centering\arraybackslash}p{1.25cm}}|*{3}{>{\centering\arraybackslash}p{1.25cm}}|*{3}{>{\centering\arraybackslash}p{1.25cm}}}
\cline{1-10} 
\multicolumn{10}{c}{ \textbf{BASELINES}}\\
\multicolumn{1}{c}{} & \multicolumn{3}{c}{FastText} & \multicolumn{3}{c}{BERT a-b non-tuned} & \multicolumn{3}{c}{BERT a-c non-tuned} \\
 Category& Overall & Pos. & Neg.  & Overall &Pos. & Neg.  & Overall & Pos. & Neg.\\
\cline{1-10} 
OVERALL & 0.51 & 0.02 & 1.00 & 0.52 & 0.79 & 0.24 & 0.49 & 0.56 & 0.42 \\
\cline{1-10} 

SAT & 0.50 & 0.01 & 1.00 & 0.52 & 0.66 & 0.39 & 0.47 & 0.40 & 0.54 \\
U2 & 0.51 & 0.02 & 1.00 & 0.51 & 0.69 & 0.33 & 0.48 & 0.54 & 0.42 \\
U4 & 0.50 & 0.01 & 1.00 & 0.49 & 0.68 & 0.30 & 0.50 & 0.55 & 0.44 \\
SCAN & 0.51 & 0.03 & 1.00 & 0.52 & 0.87 & 0.17 & 0.49 & 0.60 & 0.38 \\
SCAN - \emph{Science} & 0.57 & 0.13 & 1.00 & 0.51 & 0.84 & 0.19 & 0.43 & 0.51 & 0.35 \\
SCAN - \emph{Metaphor} & 0.50 & 0.01 & 1.00 & 0.52 & 0.88 & 0.16 & 0.50 & 0.62 & 0.39 \\

\end{tabular*}

\begin{tabular*}{\textwidth}{@{\extracolsep{\fill}}c||*{3}{>{\centering\arraybackslash}p{1.25cm}}|*{3}{>{\centering\arraybackslash}p{1.25cm}}|*{3}{>{\centering\arraybackslash}p{1.25cm}}}
\cline{1-10} 
\multicolumn{10}{c}{ \textbf{TRAINED MODELS}}\\
\multicolumn{1}{c}{}& \multicolumn{3}{c}{Simple Classification} & \multicolumn{3}{c}{BERT a-b} & \multicolumn{3}{c}{BERT a-c}  \\
 Category& Overall & Pos. & Neg.  & Overall &Pos. & Neg.  & Overall & Pos. & Neg.\\
\cline{1-10} 
OVERALL & 0.66 & 0.73 & 0.58 & \textbf{0.72}\textuparrow & \textbf{0.71}\textdownarrow & \textbf{0.72}\textuparrow & \textbf{0.52}\textuparrow & \textbf{0.67}\textuparrow & \textbf{0.37}\textdownarrow \\
\cline{1-10} 

SAT & 0.57 & 0.78 & 0.37 & \textbf{0.59}\textuparrow & 0.58\textdownarrow & \textbf{0.65}\textuparrow & \textbf{0.53}\textuparrow & 0.67\textuparrow & \textbf{0.38}\textdownarrow \\
U2 & 0.57 & 0.66 & 0.48 & 0.58\textuparrow & \textbf{0.56}\textdownarrow & \textbf{0.59}\textuparrow & 0.51\textuparrow & \textbf{0.62}\textuparrow & 0.40\textdownarrow \\
U4 & 0.60 & 0.70 & 0.50 & \textbf{0.56}\textuparrow & \textbf{0.55}\textdownarrow & \textbf{0.61}\textuparrow & 0.54\textuparrow & \textbf{0.71}\textuparrow & \textbf{0.36}\textdownarrow \\
SCAN & 0.71 & 0.74 & 0.68 & \textbf{0.82}\textuparrow & \textbf{0.77}\textdownarrow & \textbf{0.86}\textuparrow & 0.52\textuparrow & \textbf{0.67}\textuparrow & 0.37\textdownarrow \\
SCAN - \emph{Science} & 0.77 & 0.85 & 0.68 & \textbf{0.87}\textuparrow & 0.87\textuparrow & \textbf{0.88}\textuparrow & \textbf{0.50}\textuparrow & 0.57\textuparrow & 0.43\textuparrow \\
SCAN - \emph{Metaphor} & 0.69 & 0.71 & 0.68 & \textbf{0.80}\textuparrow & \textbf{0.75}\textdownarrow & \textbf{0.85}\textuparrow & 0.52\textuparrow & \textbf{0.69}\textuparrow & 0.35\textdownarrow \\
\cline{1-10} 
\end{tabular*}
\end{centering}

\caption{Average Accuracy on Analogy Classification Task. Arrows next to \emph{BERT a-b} are labeling whether accuracy went up\textuparrow, down\textdownarrow, or stayed the same\textrightarrow, as compared to \emph{BERT a-b non-tuned}. \emph{BERT a-c} is compared to \emph{BERT a-c non-tuned}. Boldface indicates a statistically significant change (p<0.05) using a z-test for proportions.}
\label{tab:table1}
\end{table*}

\subsubsection{Baselines: No fine-tuning and FastText}

We have three non-finetuned baselines, which we use to test the impact of our model training. The first is FastText\footnote{gensim library: https://radimrehurek.com/gensim/index.html}, which was used because of its large vocabulary and ability to handle out of vocabulary words \cite{bojanowski2017enriching}. Given an analogy pair \(a:b::c:d\), the cosine similarity \emph{cos((a-b),(c-d))} was calculated. In addition to FastText, we also used two pretrained SBERT baselines: \emph{BERT a-b non-tuned} and \emph{BERT a-c non-tuned}. These last two use the same cosine similarity measures as \emph{BERT a-b} and \emph{BERT a-c}, which will allow a more direct comparison to evaluate learning capabilities.

\subsection{Datasets}
For fine-tuning, we use a combination of four datasets (described below). This data was balanced, with half of the data points being true analogies and half false. We generated one negative analogy from every positive analogy, where for every true \(a:b::c:d\), a negative analogy \(a:b::c':d'\) was created, which resultsed in a total of 4930 analogies. The choice of \emph{c',d'} depended on the dataset and is described in their respective subsections.  Examples of analogies contained in the datasets used, as well as other characteristics of the data, are in Table ~\ref{tab:table7} in appendix \ref{sec:appendixa}. The data was split into ten parts (each equal to $\approx$10\% of the data). When creating
each test set, we randomly selected positive analogies, and included their negative counterpart so that the model would not see either. Each model was trained ten times with one of the portions held out. Results shown are averaged across all ten runs.

\subsubsection{SAT Dataset}
The first dataset is composed of 374 analogies taken from a portion of the SAT college entrance exam in the US (this section has since been discontinued), and has been used in several NLP publications \cite{turney2003learning, Ushio_2021, ushio2022bert}. The original format was multiple choice, where, given a pair of words, the student had to choose another pair of words out of five provided that best formed an analogy with the given pair. Each question has one valid pair. One incorrect pair from the remaining four incorrect choices was chosen for each analogy to create negative samples, creating a total of 748 SAT analogies. One beneficial quality about these negative edges is that the questions were originally developed to be challenging to humans, therefore the incorrect option is not a pair of two random entities, but instead two entities that likely were chosen to be tricky.

\subsubsection{U2 and U4}
These analogies come from an English language learning website\footnote{https://englishforeveryone.org/Topics/Analogies.html}, used in some previous NLP analogy publications \cite{Ushio_2021,ushio2022bert,boteanu2015solving}. They were made for approximately ages 9-18, therefore comprising of a range of difficulty for humans. These questions were originally formatted as multiple choice as well, so negative instances were created in the same way as with the SAT data. The U2 dataset is a subset of the U4 dataset, so we removed all analogies that were present in the U2 from the U4. These two datasets contributed a total of 1208 analogies.

\subsubsection{Scientific and Creative Analogy dataset (SCAN)}
The final dataset used in training is the Scientific and Creative Analogy dataset (SCAN), and is made up of analogies found in science, as well as commonly used metaphors in English literature \cite{czinczoll2022scientific}. Instead of forming pairs of pairs (with 4 entities), each analogy pair is composed of multi-entity relations. For example, in the analogy comparing the \emph{solar system} to an \emph{atom}, \emph{solar system} includes the entities, \(sun, planet, gravity\), while \emph{atom} includes the analogous entities \(nucleus,electron,electromagnetism\). Each analogy pair provides \(C(n,2)\) total analogies in the format \(a:b::c:d\), where \emph{n} is the number of entities in a topic. So, for example, in an analogy where each topic in a pair contains four entities, there are \(C(4,2) = 6\) total analogies. While analogies that make up the evaluation test were not seen during training, the entities in the pairs that make up the evaluation analogies may have been seen, given the multi-entity nature of this dataset. This allows us to test the model's ability to learn to infer analogical relations when it is given other relations the entities do and do not have. Negative edges were created by randomly shuffling the \emph{c,d} terms in the dataset. Given that the analogies were formed from combinations, random shuffling may accidentally result in a true analogy. All negative analogies were checked to make sure that they were not actually present in the positive analogy group. This created a total of 3102 analogies from this dataset - this represents about 63\% of the total data.

\begin{table*}[t!]
\setlength\tabcolsep{3pt}
 \begin{centering}


\begin{tabular*}{\textwidth}{c||*{3}{>{\centering\arraybackslash}p{2cm}}||*{3}{>{\centering\arraybackslash}p{1.8cm}}}
\cline{1-7} 
\multicolumn{1}{c}{} & \multicolumn{3}{c}{\textbf{BASELINES}} & \multicolumn{3}{c}{\textbf{FINE-TUNED MODELS}}\\
& FastText & BERT& BERT& Simple & BERT&   BERT\\

Category& &  a-b non-tuned &a-c non-tuned &Classification &a-b & a-c\\
\cline{1-7} 
OVERALL & 0.81 & 0.69 & 0.46 & 0.54 & \textbf{0.84}\textuparrow & \textbf{0.55}\textuparrow \\
SAT & 0.87 & 0.63 & 0.47 & 0.52 & \textbf{0.82}\textuparrow & \textbf{0.55}\textuparrow \\
U2 & 0.76 & 0.71 & 0.49 & 0.52 & \textbf{0.83}\textuparrow & \textbf{0.59}\textuparrow \\
U4 & 0.75 & 0.71 & 0.43 & 0.56 & \textbf{0.84}\textuparrow & \textbf{0.56}\textuparrow \\
SCAN & 0.82 & 0.70 & 0.46 & 0.55 & \textbf{0.85}\textuparrow & \textbf{0.54}\textuparrow \\
SCAN - \emph{Science} & 0.91 & 0.69 & 0.46 & 0.56 & \textbf{0.81}\textuparrow & \textbf{0.58}\textuparrow \\
SCAN - \emph{Metaphor} & 0.80 & 0.70 & 0.46 & 0.55 & \textbf{0.86}\textuparrow & \textbf{0.53}\textuparrow \\
\cline{1-7} 

\end{tabular*}

\caption{Average Accuracy on the Analogy Ranking Task. Arrows next to \emph{BERT a-b} are labeling whether accuracy went up\textuparrow, down\textdownarrow, or stayed the same\textrightarrow, as compared to \emph{BERT a-b non-tuned}. \emph{BERT a-c} is compared to \emph{BERT a-c non-tuned}. Boldface indicates a statistically significant change (p<0.05) using a z-test for proportions.}
\label{tab:table2}

  \end{centering}
\end{table*}

\subsubsection{Human Baseline Comparison: Distractor Dataset}
These analogies were compiled by researchers in a university psychology department, where they tested whether semantic similarity affected an adult human's ability to correctly solve analogies \cite{jones2022differential}. This dataset was not used in our model training, and recently has been used to probe large language model for analogical reasoning ability \cite{webb2023emergent}. In the original paper, the human subjects are presented with an incomplete analogy \(a:b::c:?\), where they must choose between two options for the \emph{d} term. There are two levels of semantic similarity the authors explored. First they test human's abilities to solve analogies with regards to how related the \emph{c,d} term is to the \emph{a,b} term. Analogies are grouped into near analogies, where the \emph{a,b} entities are semantically similar to the \emph{c,d} entities, and far analogies, where the \emph{a,b} entities are not semantically similar to the \emph{c,d} pair. Then within each of these groups, they come up with two incorrect \emph{d} options, which they refer to as distractors (the incorrect choices for \emph{d}). One of the incorrect \emph{d} entities is more semantically similar to the \emph{c} term than the true \emph{d} term is to the \emph{c} term, which they refer to as a high distractor salience. For example, a true analogy they use is \(tarantula:spider::bee:insect\). They replace \emph{insect} with \emph{hive} as it is more related to \emph{bee}. They measured semantic distance using LSA. The second incorrect \emph{d} term that was chosen was less semantically similar (ex: replacing \emph{insect} with \emph{yellow}), which they refer to as low distractor salience. They also test three types of analogical relations: categorical, compositional and causal. Definitions and examples of these relations can be found in \citet{jones2022differential}.

\subsubsection{External Tasks: Semantic Similarity}
In order to see if our training scheme affects performance on external tasks, as can be the case with catastrophic forgetting, we test our models on three word-level, non-contextual semantic similarity datasets: SimLex-999, MEN, and WordSim353 (WS353) \cite{GoodfellowMDCB13, articlekirk, hill-etal-2015-simlex,bruni,finkelstein2001placing, kemker2018measuring}\footnote{ We used some code from https://github.com/kudkudak/word-embeddings-benchmarks}. All these datasets contain words pairs with a similarity measure, however  \citet{hill-etal-2015-simlex} details some key features of how these dataset differ when they introduced SimLex-999; namely that both the MEN and WS353 tended to measure word relatedness/association as opposed to word similarity (not that these are mutually exclusive), and MEN's tendency to focus on less abstract concepts, such as nouns.
The similarity measure within these datasets range from 0 to 10, while we use cosine similarity (details described in the next section), which gives a similarity measure in the range -1 to 1. We chose these tasks because it is a word level similarity task, which is related to our analogy task. Ideally the performance on these tasks would improve with our training, or minimally not decrease.

\begin{table*}[h]
\setlength\tabcolsep{2pt}
\begin{centering}
\begin{tabular*}{\textwidth}{@{\extracolsep{5pt}}c|*{2}{>{\centering\arraybackslash}p{2.65cm}}||c||*{2}{>{\centering\arraybackslash}p{2.65cm}}|c}
\cline{1-7} 
& \multicolumn{2}{c||}{\textbf{Before finetuning}}&& \multicolumn{2}{c|}{\textbf{After finetuning}}&\\

\cline{1-7} 

 &  Model Positive& Model Negative    &  &  Model Positive& Model Negative    &\\
\cline{1-7} 
 True Positive &205376&108669&  &185999&184357&185524\\
  True Negative &207319& 141762&&168606& 200765&191815\\
\cline{1-7} 
   &206329& 126385&&181109& 196079&\\
  \cline{1-7} 

\end{tabular*}
\caption{Average \# of times entities seen in pre-training data by true label and model guess before(B)/after(A) fine tuning (\emph{BERT a-b})}
\label{tab:table3}
\end{centering}

\end{table*}

\begin{table*}[h]

\begin{centering}
\begin{tabular*}{\textwidth}{@{\extracolsep{1pt}}c|cc|c||cc|c}
\cline{1-7} 
& \multicolumn{2}{c|}{\textbf{Before finetuning}}&& \multicolumn{2}{c|}{\textbf{After finetuning}}&\\

\cline{1-7} 

 &  No OOV entity& 1+ OOV entity & Total &  No OOV entity& 1+ OOV entity   &Total\\
\cline{1-7} 
 True Positive &0.98&0.54&0.79  &0.75&0.65&0.71\\
  True Negative &0.03&0.53&0.24 &0.73&0.72&0.72\\
\cline{1-7} 
   &0.50&0.53&  &0.74&0.69&\\
  \cline{1-7} 

\end{tabular*}
\caption{Accuracy among analogies with no OOV entities and those with at least one before/after fine tuning (\emph{BERT a-b}). No OOV n=2889, OOV n= 2041}
\label{tab:table4}
\end{centering}
\end{table*}

\begin{table*}[h]
\setlength\tabcolsep{3pt}
\begin{centering}

\begin{tabular*}{\textwidth}{@{\extracolsep{\fill}}*{1}{>{\centering\arraybackslash}p{1.75cm}}*{1}{>{\centering\arraybackslash}p{2.25cm}}||*{3}{>{\centering\arraybackslash}p{1cm}}|*{3}{>{\centering\arraybackslash}p{1cm}}|*{3}{>{\centering\arraybackslash}p{1cm}}}
\cline{1-11} 
\multicolumn{11}{c}{ \textbf{BASELINES}}\\

\multicolumn{2}{c}{} &  \multicolumn{3}{c}{FastText} & \multicolumn{3}{c}{BERT a-b non-tuned} & \multicolumn{3}{c}{BERT a-c non-tuned} \\
\cline{3-11} 
\multicolumn{1}{c}{Semantic } &
\multicolumn{1}{c}{Relation}& \multicolumn{3}{c}{Distractor Salience} & \multicolumn{3}{c}{Distractor Salience} & \multicolumn{3}{c}{Distractor Salience} \\

 Distance&Type& Overall & High &Low& Overall &  High &Low& Overall &  High &Low\\
\cline{1-11} 
 & OVERALL & 0.70 & 0.63 & 0.77 & 0.53 & 0.52 & 0.53 & 0.34 & 0.27 & 0.42 \\
 \cline{1-11} 

Near & Overall & 0.82 & 0.77 & 0.87 & 0.53 & 0.50 & 0.57 & 0.32 & 0.27 & 0.37 \\
 & Categorical & 0.75 & 0.70 & 0.80 & 0.55 & 0.50 & 0.60 & 0.15 & 0.20 & 0.10 \\
 & Causal & 0.70 & 0.60 & 0.80 & 0.55 & 0.50 & 0.60 & 0.35 & 0.20 & 0.50 \\
 & Compositional & 1.00 & 1.00 & 1.00 & 0.50 & 0.50 & 0.50 & 0.45 & 0.40 & 0.50 \\
Far & Overall & 0.58 & 0.50 & 0.67 & 0.52 & 0.53 & 0.50 & 0.37 & 0.27 & 0.47 \\
 & Categorical & 0.75 & 0.70 & 0.80 & 0.65 & 0.60 & 0.70 & 0.30 & 0.20 & 0.40 \\
 & Causal & 0.55 & 0.50 & 0.60 & 0.45 & 0.50 & 0.40 & 0.30 & 0.30 & 0.30 \\
 & Compositional & 0.45 & 0.30 & 0.60 & 0.45 & 0.50 & 0.40 & 0.50 & 0.30 & 0.70 \\

\end{tabular*}

\begin{tabular*}{\textwidth}{@{\extracolsep{\fill}}*{1}{>{\centering\arraybackslash}p{1.75cm}}*{1}{>{\centering\arraybackslash}p{2.25cm}}||*{3}{>{\centering\arraybackslash}p{1cm}}|*{3}{>{\centering\arraybackslash}p{1cm}}|*{3}{>{\centering\arraybackslash}p{1cm}}}
\cline{1-11} 
\multicolumn{11}{c}{ \textbf{TRAINED MODELS}}\\
\multicolumn{2}{c}{} & \multicolumn{3}{c}{Simple} & \multicolumn{3}{c}{BERT a-b} & \multicolumn{3}{c}{BERT a-c}  \\
\cline{3-11} 
\multicolumn{1}{c}{} &
\multicolumn{1}{c}{}& \multicolumn{3}{c}{Distractor Salience} & \multicolumn{3}{c}{Distractor Salience} & \multicolumn{3}{c}{Distractor Salience} \\

&&  Overall & High &Low&  Overall & High &Low& Overall & High &Low\\
\cline{1-11} 
  & OVERALL & 0.52 & 0.51 & 0.54 & \textbf{0.69}\textuparrow & 0.68\textuparrow & 0.70\textuparrow & 0.45\textuparrow & 0.35\textuparrow & 0.54\textuparrow \\
  
  \cline{1-11} 
Near & Overall & 0.63 & 0.56 & 0.70 & \textbf{0.75}\textuparrow & 0.71\textuparrow & 0.79\textuparrow & 0.45\textuparrow & 0.37\textuparrow & 0.52\textuparrow \\
 & Categorical & 0.64 & 0.57 & 0.70 & 0.73\textuparrow & 0.71\textuparrow & 0.75\textuparrow & 0.50\textuparrow & 0.44\textuparrow & \textbf{0.55}\textuparrow \\
 & Causal & 0.63 & 0.58 & 0.67 & 0.72\textuparrow & 0.65\textuparrow & 0.78\textuparrow & 0.48\textuparrow & 0.41\textuparrow & 0.54\textuparrow \\
 & Compositional & 0.63 & 0.53 & 0.73 & \textbf{0.82}\textuparrow & 0.78\textuparrow & 0.85\textuparrow & 0.37\textuparrow & 0.26\textdownarrow & 0.47\textuparrow \\
Far & Overall & 0.42 & 0.46 & 0.37 & 0.62\textuparrow & 0.64\textuparrow & 0.61\textuparrow & 0.45\textuparrow & 0.33\textuparrow & 0.56\textuparrow \\
 & Categorical & 0.46 & 0.55 & 0.37 & 0.67\textuparrow & 0.73\textuparrow & 0.60\textdownarrow & 0.46\textuparrow & 0.30\textuparrow & 0.62\textuparrow \\
 & Causal & 0.36 & 0.38 & 0.34 & 0.65\textuparrow & 0.66\textuparrow & 0.63\textuparrow & 0.45\textuparrow & 0.38\textuparrow & 0.51\textuparrow \\
 & Compositional & 0.44 & 0.46 & 0.41 & 0.56\textuparrow & 0.53\textuparrow & 0.59\textuparrow & 0.43\textdownarrow & 0.32\textuparrow & 0.54\textdownarrow \\ 
 \cline{1-11}

\end{tabular*}
\caption{Average Accuracy on Distractor Dataset. Arrows next to \emph{BERT a-b} are labeling whether accuracy went up\textuparrow, down\textdownarrow, or stayed the same\textrightarrow, as compared to \emph{BERT a-b non-tuned}. \emph{BERT a-c} is compared to \emph{BERT a-c non-tuned}. Boldface indicates a statistically significant change (p<0.05) using a z-test for proportions.}
\label{tab:table5}
\end{centering}
\end{table*}

\section{Results} 

\subsection{Proportional Analogies as a Learnable Objective for Neural Networks}

Table~\ref{tab:table1} shows accuracy on classifying the testset with both the baselines and trained models. FastText has a tendency to label all analogies as negative given the cosine similarity measure, while BERT models have a tendency to classify all analogies as positive. This is perhaps unsurprising, as it has been demonstrated that word embeddings exhibit anisotropy, and that anisotropy is higher with contextual word embeddings, resulting in any two random words having high cosine similarity to each other, perhaps translating into the distances between words being similar to each other \cite{ethayarajh2019contextual}. \emph{BERT a-b} seems to be less likely to be biased towards one label as compared to the other baselines, with the relatively large SCAN dataset having the greatest tendency to be classified as positive. 

The \emph{a-b} training scheme improved overall accuracy on analogy classification over the previously discussed baseline, with most positive changes in accuracy being statistically significant. The largest gains were with the SCAN dataset, mostly due to an increased ability to correctly classify negative analogies. Performance was generally better with the metaphor analogies than science, with the \emph{a-b} model reaching 0.87 accuracy overall. \citet{czinczoll2022scientific} found that models performed better on science analogies than on metaphor analogies, which they attributed to metaphors being more abstract. As mentioned before, while the model would have never seen the examples from the evaluation set, it would have seen the entities in the pairs that make up the samples in the evaluation set as parts of other analogies. There was improvement on the other datasets with the \emph{a-b} model, however the overall improvements were less dramatic, with accuracy +0.07 when compared to \emph{BERT a-b non-tuned}. We cannot directly compare our results to \citet{czinczoll2022scientific} and \citet{ushio2022bert}, as \citet{czinczoll2022scientific} does not use a classification or ranking multiple choice task while \citet{ushio2022bert} used the entire list of negative analogies for the ranking task. Moreover, the main goals of these papers were not to test training schemes.

The fine-tuned and non-fine-tuned models were generally able to perform an analogy ranking task better than the classification task, as shown in Table ~\ref{tab:table2}. The performance of each model between SCAN and the other datasets was less variable with the ranking task as opposed to the classification task. Again, the fine-tuned models outperformed their respective baselines with statistically significant improvements, the improvements being much greater with the \emph{a-b} scheme.

\subsection{Exploring Accuracy in Relation to Word Frequency and Subwords}

Inspired by \citet{zhou2022problems, zhou2022richer}, we explored whether there were any trends in classification associated with entity frequency in BERT's pre-training data, as well as subword tokenization. They found that cosine similarity between BERT embeddings tends to be under-estimated among frequent words \cite{zhou2022problems}. They also found that countries who were out-of vocabulary (OOV) in BERT's vocabulary were more likely to be judged as similar to other countries, and being OOV was related to being mentioned less in BERT's pre-training data \cite{zhou2022richer}. Keep in mind that we classified analogies using the cosine similarity between the distances between entities, and not the cosine between the entities themselves, which differentiates our results from theirs.

In order to approximate word frequency in the training data for our experiments, we use the estimates released by \cite{zhou2022problems}. Like \citet{zhou2022richer} found, the less common a word in the training data is, the more likely it is to be out-of vocabulary (OOV) (Figure ~\ref{fig:Figure_1} in appendix A). Words tokenized into two or more subwords have generally been seen <10,000 times in the training data. Table \ref{tab:table8} in appendix A shows the percent of each dataset that contains OOV words, as well as average times an entity is seen in the training data. The SCAN dataset contained < 10\% OOV entities, while the SAT dataset contained almost 30\% OOV words. Entities in the SCAN were seen on average twice as much in the pre-training data as entities in the SAT dataset.

Table \ref{tab:table3} shows average word frequency by true and predicted label, while Table \ref{tab:table4} shows classification accuracy by whether an analogy had at least one OOV entity. The entities contained in false analogies tended to be observed in the pre-training data more frequently than those in true analogies. However, analogies predicted as being true analogies contained entities that were seen a little over 60\% more on average than those contained in analogies predicted as false before training. Additionally, analogies that contained no OOV entities were almost always predicted as true before training (Table \ref{tab:table4}). After training, the average frequency among predicted labels closely matches that among the true labels, and accuracy improved greatly among negative analogies with no OOV entities, as well as among analogies with OOV entities. It appears that before fine-tuning, the model overestimated the similarity in relations between analogy pairs with in-vocabulary words, and a bulk of the learning affected the ability to correctly identify lack of analogy. Similar trends can be seen when looking within each dataset.

\subsection{Neural Networks as Compared to Humans}

Table ~\ref{tab:table5} shows the results of testing our methods on an unseen testset that was previously tested on college students in the US by \citet{jones2022differential}. As a summary of what the original paper found - humans overall did well on solving these analogies ($\approx$84\% accuracy overall). They found that humans were better at solving near analogies than far analogies, and that humans had a harder time correctly choosing \emph{d} then when there was high distractor salience as compared to a low distractor salience. When looking at relation categories, human performance was highest on the categorical analogies, and lowest on the causal analogies. To see results in detail please refer to \citet{jones2022differential}.

\begin{table*}[h]
\setlength\tabcolsep{4pt}
\begin{center}

\begin{tabular*}{\textwidth}{@{\extracolsep{5pt}}cccccc||cc||*{3}{>{\centering\arraybackslash}p{1.75cm}}}
\cline{6-11} 
\multicolumn{6}{c}{} & \multicolumn{2}{c}{Baselines} & \multicolumn{3}{c}{Fine-tuned Models}\\

&&&&&& FastText & BERT& Simple & BERT & BERT \\
 &&&&&Tasks& &   non-tuned&  Classification &  a-b&  a-c  \\
 
\cline{6-11} 
&&&&&SimLex-999 & 0.44 & 0.17 &0.21 & \textbf{0.33}\textuparrow &0.19\textuparrow\\
&&&&&MEN & 0.81 & 0.27 & 0.27& \textbf{0.31}\textuparrow& \textbf{0.33}\textuparrow\\
&&&&&WS535 & 0.69 & 0.27 &0.27 &0.27\textrightarrow &0.30\textuparrow\\
\cline{6-11}

\end{tabular*}
\caption{Average Performance on Several Semantic Similarity Tasks. Arrows next to \emph{BERT a-b} are labeling whether accuracy went up\textuparrow, down\textdownarrow, or stayed the same\textrightarrow, as compared to \emph{BERT non-tuned}. \emph{BERT a-c} is also compared to \emph{BERT non-tuned}, since this task compares the similarity between two words, not between the differences between two words. Boldface indicates a statistically significant change (p<0.05) using a z-test for proportions.}
\label{tab:table6}
\end{center}
\end{table*}

 In our experiments, the best performing model was \emph{BERT a-b}, with a 0.69 overall accuracy, up from 0.53 with \emph{BERT a-b non-tuned}, and $\approx$0.15 worse than human performance. Accuracy for \emph{BERT a-b} mostly increased with training over the baseline, however most increases among subgroups were not statistically significant, although notably the sample size was small. When looking at subgroups, the same trends observed in humans were not present, nor were there any obvious trends among subgroups between the models, with the exception that near analogies were easier to solve than far analogies for our best model.

\subsection{Performance on External Tasks}

Finally, we tested on an external task to find out whether fine-tuning on the task of analogical reasoning might have a (negative) effect on semantic similarity tasks. Table ~\ref{tab:table6} shows the Spearman's rank-order correlation coefficient for the three Semantic Similarity tasks. \emph{BERT a-b} improved over \emph{BERT non-tuned}, showing that training actually improved performance, even if performance is still overall low compared to FastText. FastText outperformed all transformer models on the external tasks, which is unsurprising. \citet{ethayarajh2019contextual} found that FastText and embeddings from lower layers of BERT outperformed final layer hidden representations from  BERT, although they used the first principal component of the embeddings so the results are not directly  comparable. Given the tasks are non-contextual, perhaps the contextual nature of BERT that allows it to perform well on certain tasks hinders it in others. 
Interestingly \emph{BERT a-b} performed better on the SimLex-999 task than the other two tasks, unlike the baseline models presented. \citet{hill-etal-2015-simlex} had found the SimLex-999 task was harder for neural embeddings to solve than MEN or WS353, which they attributed to these models being better at identifying word association than similarity. However they did not test BERT-like models.

\section{Conclusion, Limitations and Future Work}

In this paper, we aimed to move from testing relatively simple analogical relations in pretrained language models to testing the ability to learn more complex relations used for testing human analogical reasoning, with a tailored training objective. We found that overall, analogies are something that can be learned. We reach an accuracy of 0.69 up from 0.53 while being 0.15 below the human upper bound, on an unseen test set constructed to test human analogical reasoning. Lastly, we find that fine-tuning models with certain training objectives generally does not deteriorate their performance on external, but related tasks. In fact, on some tasks we observed improved accuracy.

Our experiments involve several limitations. For one, the dataset is small, making claims that analogical reasoning is something that for sure can or cannot be learned with language models is not possible.
Another important consideration is that analogies are permutable \cite{ushio2022bert, marquer2022transferring}. Given an analogy (1) \(a:b::c:d\), the following analogies also hold: (2) \(b:a::d:c\), (3) \(c:d::a:b\), (4) \(d:c::b:a\), (5) \(a:c::b:d\), (6) \(c:a::d:b\), (7) \(b:d::a:c\), (8) \(d:b::c:a\). Our \emph{a-b} models account for 1-4, while our \emph{a-c} models account for 5-8. These permutations are not without criticism - specifically (5) \(a:c::b:d\) and all its derivatives (6-8) \cite{marquer2022transferring}. Consider the analogy \((electron : nucleus :: planet : sun)\). In our \emph{a-b} models, we are making the assumption \emph{cos((a-b),(c-d))}. In natural language, the relation on either side could be verbalized as \emph{revolves around}. In the measure \emph{cos((a-c),(b-d))}, there is no verbalizer that can describe both the \emph{a,c} and \emph{b,d} equivalently. The question is if that corresponds to no vector transformation that is equivalent between the two pairs. Lastly, \citet{czinczoll2022scientific} mentioned some of the metaphorical analogies contained antiquated gender roles, which could be potentially harmful.

One direction for future work is to address limitations of word embeddings or their representation power, such as the anisotropy that exists among contextual word embeddings \cite{ethayarajh2019contextual}. Perhaps a distance measure between two entities that is not subtraction would be a better way to represent their relation. Additionally, since prior work has suggested that analogies where the entities are close to each other in space are generally easier to solve with the vector offset method, perhaps focusing on incorporating knowledge regarding semantic distance between entities during training would be helpful \cite{rogers-etal-2017-many}. We would also like to explore whether augmenting LM training with analogy learning for other common NLP benchmarks affects performance on these benchmarks.

\section*{Acknowledgements}

We are grateful to the Swiss National Science
Foundation (\(grant~205121\_207437: C-LING\))  for
funding this work. We also thank members of the
Idiap NLU-CCL group, the NLP lab at EPFL, the Programme group Psychological Methods of the University of Amsterdam for helpful discussions,
and the anonymous reviewers
for their fruitful comments and suggestions.

\bibliography{bibliography}
\bibliographystyle{acl_natbib}
\clearpage
\onecolumn
\appendix

\section{Characteristics of our Dataset}
\label{sec:appendixa}

\begin{figure}[h]
    \centering
    \includegraphics[scale=0.6]{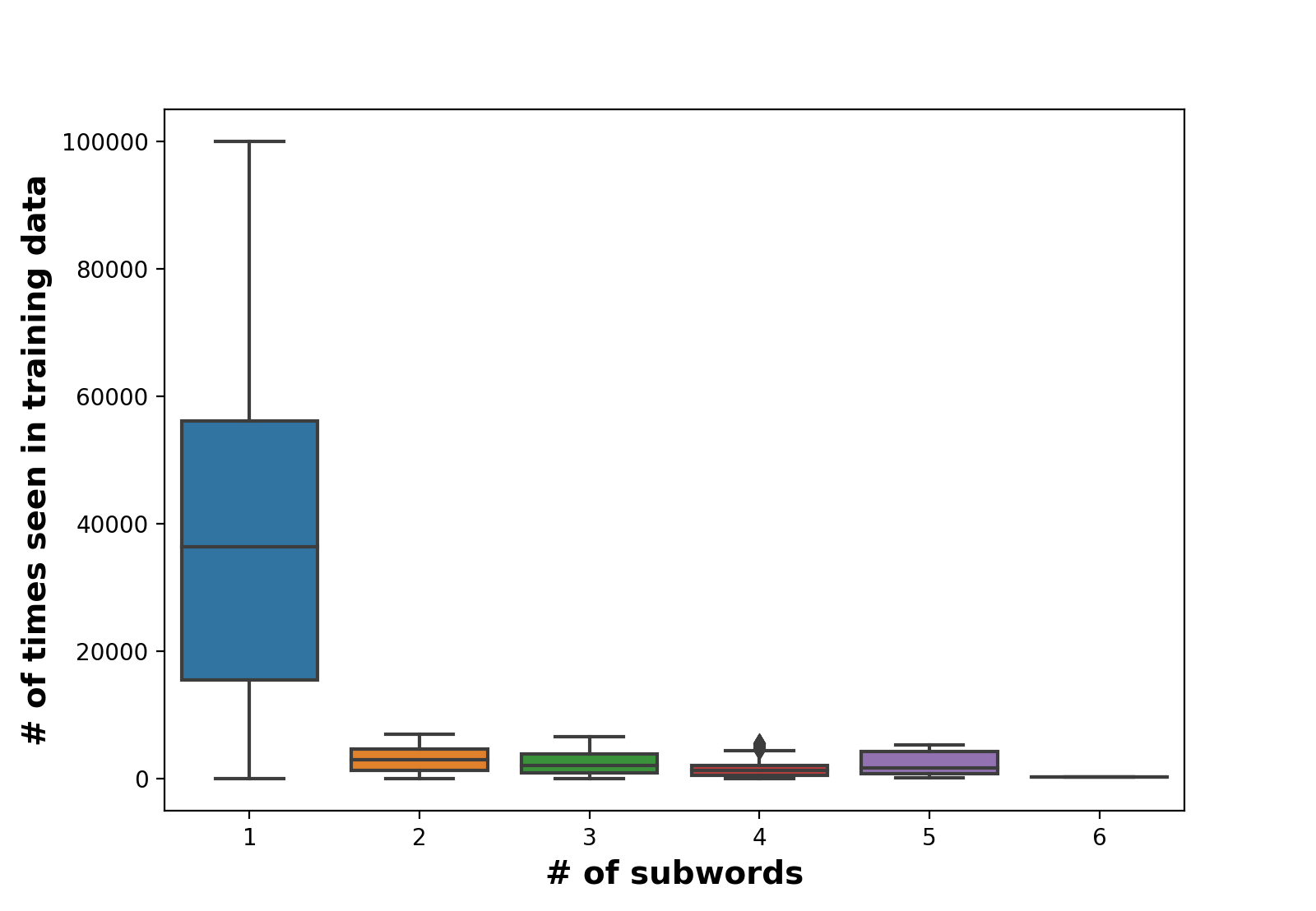}
    \caption{Distribution of estimated word frequency seen in pre-training data, by number of token per word (among words seen <100,000 times)}
    \label{fig:Figure_1}
\end{figure}

Tables \ref{tab:table7} and \ref{tab:table8} as well as Figure \ref{fig:Figure_1} characterize the training data that we used to reach a better understanding of what the model actually 'sees' during training. For explanations on how we created the training data from the original datasets we refer the reader to the section on Datasets. Table \ref{tab:table7} gives examples of analogies and their negative counterparts used in the data. Table \ref{tab:table8} shows some characteristics of the entities that make up each dataset we used for training our models. 

\begin{table*}[ht]
\setlength\tabcolsep{2pt}
\begin{centering}

\begin{tabular*}{\textwidth}{@{\extracolsep{5pt}}c|c|ccc|ccc}
\cline{1-8} 
\multicolumn{1}{c}{}& \multicolumn{1}{c}{n} & \multicolumn{3}{c}{Positive}& \multicolumn{3}{c}{Negative}   \\
\cline{1-8} 
 SCAN&2974 &\multicolumn{3}{c}{nucleus:electron::sun:planet}& \multicolumn{3}{c}{nucleus:electron::traveler:station} \\
  SAT&748 &\multicolumn{3}{c}{amalgam:metals::coalition:factions}& \multicolumn{3}{c}{amalgam:metals::car:payments}\\
   U2&504 &\multicolumn{3}{c}{permanent:temporary::skeptical:trusting}& \multicolumn{3}{c}{permanent:temporary::ordinary:plain} \\
    U4&704 &\multicolumn{3}{c}{order:chaos::unity:division}& \multicolumn{3}{c}{order:chaos::culture:feeling}  \\
\cline{1-8}

\end{tabular*}
\caption{Example of Analogies in the Datasets used for Training}
\label{tab:table7}
\end{centering}

\end{table*}

\begin{table*}[ht]
\setlength\tabcolsep{2pt}
\begin{centering}

\begin{tabular*}{\textwidth}{@{\extracolsep{30pt}}c|ccc|ccc}
\cline{1-7} 
\multicolumn{1}{c}{} & \multicolumn{3}{c}{\% OOV}& \multicolumn{3}{c}{Ave. \# of times entity seen in pre-training data}   \\
\cline{1-7} 
 SCAN &\multicolumn{3}{c}{9.4\%}& \multicolumn{3}{c}{217127} \\
  SAT &\multicolumn{3}{c}{29.4\%}& \multicolumn{3}{c}{126132}\\
   U2 &\multicolumn{3}{c}{21.6\%}& \multicolumn{3}{c}{158047} \\
    U4 &\multicolumn{3}{c}{23.8\%}& \multicolumn{3}{c}{156826}  \\
\cline{1-7} 
\end{tabular*}
\caption{Characteristics of Entities in Datasets}
\label{tab:table8}
\end{centering}
\end{table*}

\section{Accuracy in Relation to Word
Frequency and Subwords: Within Datasets}
\label{sec:appendixb}

Tables \ref{tab:table9} - \ref{tab:table16} reproduce tables \ref{tab:table3} and \ref{tab:table4} in the manuscript but for each dataset individually. Similar trends seen in the overall data are seen within each dataset, specifically the improvements among being able to classify false analogies correctly that contain no OOV entities, as well as the model becoming less biased with regards to the number of times entities were seen and the label it chose.

\begin{table*}[h]
\setlength\tabcolsep{2pt}
\begin{centering}
\caption{SCAN dataset: Average \# of times entities seen in pre-training data by true label and model guess before(B)/after(A) fine tuning (\emph{BERT a-b})}
\label{tab:table9}
\begin{tabular*}{\textwidth}{@{\extracolsep{5pt}}c|cc||c||cc|c}
\cline{1-7} 
& \multicolumn{2}{c||}{\textbf{A - SCAN}}&& \multicolumn{2}{c|}{\textbf{B - SCAN}}&\\

\cline{1-7} 

 &  Model Positive& Model Negative    &  &  Model Positive& Model Negative    &\\
\cline{1-7} 
 True Positive &225457&158858&  &209101&244098&217127\\
  True Negative &222250&191652 &&230867& 214842&217127\\
\cline{1-7} 
   &223893&177630 &&212499&221015 &\\
  \cline{1-7} 

\end{tabular*}

\end{centering}
\end{table*}

\begin{table*}[h]

\begin{centering}
\caption{SCAN dataset: Accuracy among analogies with no OOV entities and those with at least one before(A)/after(B) fine tuning (\emph{BERT a-b})}
\label{tab:table10}
\begin{tabular*}{\textwidth}{@{\extracolsep{1pt}}c|cc|c||cc|c}
\cline{1-7} 
& \multicolumn{2}{c|}{\textbf{A - SCAN}}&& \multicolumn{2}{c|}{\textbf{B - SCAN}}&\\

\cline{1-7} 

 &  No OOV entity& 1+ OOV entity & Total &  No OOV entity& 1+ OOV entity   &Total\\
\cline{1-7} 
 True Positive &0.97&0.62& 0.87 &0.77&0.83&0.77\\
  True Negative &0.03&0.47&0.17&0.78&0.92&0.86\\
\cline{1-7} 
   &0.51&0.54&  &0.80&0.85&\\
  \cline{1-7} 

\end{tabular*}
\end{centering}
\end{table*}

\begin{table*}[h]
\setlength\tabcolsep{2pt}
\begin{centering}
\caption{SAT dataset: Average \# of times entities seen in pre-training data by true label and model guess before(A)/after(B) fine tuning (\emph{BERT a-b})}
\label{tab:table11}
\begin{tabular*}{\textwidth}{@{\extracolsep{5pt}}c|cc||c||cc|c}
\cline{1-7} 
& \multicolumn{2}{c||}{\textbf{A - SAT}}&& \multicolumn{2}{c|}{\textbf{B - SAT}}&\\

\cline{1-7} 

 &  Model Positive& Model Negative    &  &  Model Positive& Model Negative    &\\
\cline{1-7} 
 True Positive &138714&83755&  &122769&114774&119904\\
  True Negative &143643&114737 &&119850& 143477&132359\\
\cline{1-7} 
   &141085&100264 &&121534& 131892&\\
  \cline{1-7} 

\end{tabular*}

\end{centering}
\end{table*}

\begin{table*}[h]

\begin{centering}
\caption{SAT dataset: Accuracy among analogies with no OOV entities and those with at least one before(A)/after(B) fine tuning (\emph{BERT a-b})}
\label{tab:table12}
\begin{tabular*}{\textwidth}{@{\extracolsep{1pt}}c|cc|c||cc|c}
\cline{1-7} 
& \multicolumn{2}{c|}{\textbf{A - SAT}}&& \multicolumn{2}{c|}{\textbf{B - SAT}}&\\

\cline{1-7} 

 &  No OOV entity& 1+ OOV entity & Total &  No OOV entity& 1+ OOV entity   &Total\\
\cline{1-7} 
 True Positive &0.97&0.53& 0.66 & 0.77&0.59&0.64\\
  True Negative &0.02&0.56& 0.39 &0.47&0.55&0.53\\
\cline{1-7} 
   &0.47&0.55&  &0.62&0.57&\\
  \cline{1-7} 

\end{tabular*}
\end{centering}
\end{table*}

\begin{table*}[h]
\setlength\tabcolsep{2pt}
\begin{centering}
\caption{U2 dataset: Average \# of times entities seen in pre-training data by true label and model guess before(A)/after(B) fine tuning (\emph{BERT a-b})}
\label{tab:table13}
\begin{tabular*}{\textwidth}{@{\extracolsep{5pt}}c|cc||c||cc|c}
\cline{1-7} 
& \multicolumn{2}{c||}{\textbf{A - U2}}&& \multicolumn{2}{c|}{\textbf{B - U2}}&\\

\cline{1-7} 

 &  Model Positive& Model Negative    &  &  Model Positive& Model Negative    &\\
\cline{1-7} 
 True Positive &187756&78472&  &178890&116153&153497\\
  True Negative &191153& 104450&&167793&158504 &162596\\
\cline{1-7} 
   &189435& 91782&&174171&140727 &\\
  \cline{1-7} 

\end{tabular*}

\end{centering}
\end{table*}

\begin{table*}[h]

\begin{centering}
\caption{U2 dataset: Accuracy among analogies with no OOV entities and those with at least one before(A)/after(B) fine tuning (\emph{BERT a-b})}
\label{tab:table14}
\begin{tabular*}{\textwidth}{@{\extracolsep{1pt}}c|cc|c||cc|c}
\cline{1-7} 
& \multicolumn{2}{c|}{\textbf{A - U2}}&& \multicolumn{2}{c|}{\textbf{B - U2}}&\\

\cline{1-7} 

 &  No OOV entity& 1+ OOV entity & Total &  No OOV entity& 1+ OOV entity   &Total\\
\cline{1-7} 
 True Positive &0.97&0.44& 0.67 & 0.68&0.52&0.60\\
  True Negative &0.03&0.62& 0.33 &0.48&0.63&0.56\\
\cline{1-7} 
   &0.49&0.53&  &0.58&0.57&\\
  \cline{1-7} 

\end{tabular*}
\end{centering}
\end{table*}

\FloatBarrier

\begin{table*}[h]
\setlength\tabcolsep{2pt}
\begin{centering}
\caption{U4 dataset: Average \# of times entities seen in pre-training data by true label and model guess before(A)/after(B) fine tuning (\emph{BERT a-b})}
\label{tab:table15}
\begin{tabular*}{\textwidth}{@{\extracolsep{5pt}}c|cc||c||cc|c}
\cline{1-7} 
& \multicolumn{2}{c||}{\textbf{A - U4}}&& \multicolumn{2}{c|}{\textbf{B - U4}}&\\

\cline{1-7} 

 &  Model Positive& Model Negative    &  &  Model Positive& Model Negative    &\\
\cline{1-7} 
 True Positive &177438&75383&  &147907&139783&144676\\
  True Negative &202324&90524 &&149768& 186713&168975\\
\cline{1-7} 
   &190086&82676 &&148732& 166372&\\
  \cline{1-7} 

\end{tabular*}

\end{centering}
\end{table*}

\begin{table*}[h]

\begin{centering}
\caption{U4 dataset: Accuracy among analogies with no OOV entities and those with at least one before(A)/after(B) fine tuning (\emph{BERT a-b})}
\label{tab:table16}
\begin{tabular*}{\textwidth}{@{\extracolsep{1pt}}c|cc|c||cc|c}
\cline{1-7} 
& \multicolumn{2}{c|}{\textbf{A - U4}}&& \multicolumn{2}{c|}{\textbf{B - U4}}&\\

\cline{1-7} 

 &  No OOV entity& 1+ OOV entity & Total &  No OOV entity& 1+ OOV entity   &Total\\
\cline{1-7} 
 True Positive &0.99&0.45& 0.68 & 0.69&0.53&0.60\\
  True Negative &0.00&0.58& 0.30 &0.46&0.58&0.52\\
\cline{1-7} 
   &0.47&0.51&  &0.57&0.55&\\
  \cline{1-7} 

\end{tabular*}
\end{centering}
\end{table*}

\FloatBarrier

\section{Results from architectures: Single layer}
\label{sec:appendixc}

We tested a model where we added a single
feed forward layer on top of a pretrained BERT
model, before the pooling layer. We kept the pretrained BERT weights frozen and trained only the
additional layer using the \emph{a-b} training scheme. Training with a single layer saw minimal change over the \emph{BERT a-b non-tuned} baseline presented in the main text.

\begin{table*}[ht]
\setlength\tabcolsep{2pt}
\begin{centering}
\caption{Single Layer BERT a-b: Accuracy on Analogy Classification and Ranking Task}
\label{tab:table17}

\begin{tabular*}{\textwidth}{c||*{3}{>{\centering\arraybackslash}p{2cm}}||*{1}{>{\centering\arraybackslash}p{3cm}}}
\cline{1-5} 
\multicolumn{1}{c}{} & \multicolumn{3}{c}{Classification}& \multicolumn{1}{c}{Ranking} \\
 Category& Overall & Positive & Negative&  \\
\cline{1-5} 

OVERALL & 0.53 & 0.72 & 0.33&0.69 \\
\cline{1-5} 

SAT & 0.53 & 0.61 & 0.45& 0.63 \\
U2 & 0.52 & 0.63 & 0.40 &0.72 \\
U4 & 0.48 & 0.62 & 0.34 &0.70\\
SCAN  & 0.54 & 0.79 & 0.28 &0.70\\
SCAN -\emph{Science}& 0.54 & 0.76 & 0.31&0.72 \\
SCAN -\emph{Metaphor}& 0.54 & 0.79 & 0.28&0.70 \\

\cline{1-5} 
\end{tabular*}
\end{centering}
\end{table*}

\begin{table*}[ht]
\setlength\tabcolsep{2pt}
\begin{centering}
\caption{Single Layer BERT a-b: Accuracy on Distractor Dataset}
\label{tab:table18}

\begin{tabular*}{\textwidth}{*{1}{>{\centering\arraybackslash}p{2cm}}*{1}{>{\centering\arraybackslash}p{3cm}}||*{3}{>{\centering\arraybackslash}p{2cm}}}
\cline{1-5} 
\\
\multicolumn{1}{c}{Semantic } &
\multicolumn{1}{c}{Relation}& \multicolumn{3}{c}{Distractor Salience}  \\

 Distance&Type& Overall & High &Low\\
\cline{1-5} 
&OVERALL & 0.54 & 0.52 & 0.56 \\
Near &Overall& 0.55 & 0.53 & 0.57 \\
&Categorical & 0.57 & 0.51 & 0.62 \\
&Causal & 0.55 & 0.55 & 0.55 \\
&Compositional & 0.54 & 0.53 & 0.55 \\
Far&Overall & 0.53 & 0.51 & 0.55 \\
&Categorical & 0.66 & 0.6 & 0.72 \\
&Causal & 0.44 & 0.41 & 0.46 \\
&Compositional & 0.49 & 0.51 & 0.46 \\

\cline{1-5}

\end{tabular*}
\end{centering}
\end{table*}

\FloatBarrier

\section{Results from other architectures: Different Models}
\label{sec:appendixd}

We ran a subset of our experiments with different BERT models: BERT-cased (110M parameters), BERT-large-uncased (340M parameters), and RoBERTa-base (125M parameters). Specifically the models we reproduced were \emph{BERT a-b non-tuned}, \emph{BERT a-b}, and \emph{Simple Classifier}. 

On the classification task, all the models saw some increase in overall accuracy as compared to the baseline, with RoBERTa achieving the highest overall accuracy (Tables \ref{tab:table19}-\ref{tab:table21}). RoBERTa slightly outperformed BERT-base on the classification and ranking tasks (Table \ref{tab:table21}). The untrained RoBERTa \emph{a-b} started out with a more extreme difference in accuracy between positive and negative samples in the classification task as compared to the \emph{BERT a-b}, and while training closed the gap between the accuracy (while improving overall accuracy), \emph{BERT a-b} was closer to achieving parity between the two groups. RoBERTa's performance did not transfer as well to the Distractor Dataset as compared to \emph{BERT a-b} (Table \ref{tab:table27}). BERT-large-uncased did not see as much improvement with fine-tuning, likely due to the model being too large to learn from the relatively small dataset (Table \ref{tab:table20},\ref{tab:table23},\ref{tab:tables26}).

\FloatBarrier

\begin{table*}[ht]
\setlength\tabcolsep{2pt}
\begin{centering}
\caption{BERT-cased: Accuracy on Analogy Classification Task}
\label{tab:table19}

\begin{tabular*}{\textwidth}{@{\extracolsep{\fill}}c||ccc|ccc|ccc}
\cline{1-10} 
\multicolumn{1}{c}{} & \multicolumn{3}{c}{BERT a-b non-tuned}& \multicolumn{3}{c}{Simple Classifier} &
\multicolumn{3}{c}{BERT a-b}  \\
\cline{1-10} 
 Category& Overall & Pos. & Neg.  & Overall &Pos. & Neg.  & Overall & Pos. & Neg.\\
\cline{1-10} 
OVERALL & 0.52 & 0.85 & 0.19 & 0.66 & 0.83 & 0.48 & 0.66 & 0.83 & 0.48 \\
\cline{1-10} 

SAT & 0.51 & 0.76 & 0.26 & 0.53 & 0.76 & 0.29 & 0.53 & 0.76 & 0.29 \\
U2 & 0.55 & 0.81 & 0.29 & 0.54 & 0.71 & 0.37 & 0.54 & 0.71 & 0.37 \\
U4 & 0.49 & 0.73 & 0.26 & 0.54 & 0.73 & 0.35 & 0.54 & 0.73 & 0.35 \\
SCAN & 0.52 & 0.9 & 0.13 & 0.74 & 0.89 & 0.58 & 0.74 & 0.89 & 0.58 \\
SCAN - \emph{Science} & 0.54 & 0.9 & 0.18 & 0.75 & 0.94 & 0.56 & 0.75 & 0.94 & 0.56 \\
SCAN - \emph{Metaphor} & 0.51 & 0.9 & 0.12 & 0.74 & 0.88 & 0.59 & 0.74 & 0.88 & 0.59 \\

\cline{1-10} 

\end{tabular*}
\end{centering}
\end{table*}

\FloatBarrier

\begin{table*}[ht]
\setlength\tabcolsep{2pt}
\begin{centering}
\caption{BERT-large: Accuracy on Analogy Classification Task}
\label{tab:table20}

\begin{tabular*}{\textwidth}{@{\extracolsep{\fill}}c||ccc|ccc|ccc}
\cline{1-10} 
\multicolumn{1}{c}{} & \multicolumn{3}{c}{BERT a-b non-tuned}& \multicolumn{3}{c}{Simple Classifier} &
\multicolumn{3}{c}{BERT a-b}  \\
\cline{1-10} 
 Category& Overall & Pos. & Neg.  & Overall &Pos. & Neg.  & Overall & Pos. & Neg.\\
\cline{1-10} 
OVERALL & 0.54 & 0.39 & 0.69 & 0.50 & 1.00 & 0.00 & 0.59 & 0.47 & 0.71 \\
\cline{1-10} 

SAT & 0.52 & 0.30 & 0.74 & 0.50 & 1.00 & 0.00 & 0.54 & 0.45 & 0.64 \\
U2 & 0.50 & 0.34 & 0.65 & 0.50 & 1.00 & 0.00 & 0.53 & 0.41 & 0.65 \\
U4 & 0.49 & 0.32 & 0.65 & 0.50 & 1.00 & 0.00 & 0.52 & 0.42 & 0.63 \\
SCAN & 0.56 & 0.44 & 0.69 & 0.50 & 1.00 & 0.00 & 0.63 & 0.49 & 0.76 \\
SCAN - \emph{Science} & 0.62 & 0.60 & 0.64 & 0.50 & 1.00 & 0.00 & 0.68 & 0.57 & 0.80 \\
SCAN - \emph{Metaphor} & 0.55 & 0.40 & 0.70 & 0.50 & 1.00 & 0.00 & 0.61 & 0.48 & 0.75 \\

\cline{1-10}

\end{tabular*}

\end{centering}

\end{table*}

\FloatBarrier

\begin{table*}[hbt!]
\setlength\tabcolsep{2pt}
\begin{centering}
\caption{RoBERTa: Accuracy on Analogy Classification Task}
\label{tab:table21}

\begin{tabular*}{\textwidth}{@{\extracolsep{\fill}}c||ccc|ccc|ccc}
\cline{1-10} 
\multicolumn{1}{c}{} & \multicolumn{3}{c}{RoBERTa a-b non-tuned}& \multicolumn{3}{c}{Simple Classifier} &
\multicolumn{3}{c}{RoBERTa a-b}  \\
\cline{1-10} 
 Category& Overall & Pos. & Neg.  & Overall &Pos. & Neg.  & Overall & Pos. & Neg.\\
 \cline{1-10} 

OVERALL & 0.51 & 0.92 & 0.10 & 0.55 & 0.52 & 0.59 & 0.73 & 0.84 & 0.62 \\
\cline{1-10} 

SAT & 0.51 & 0.88 & 0.14 & 0.52 & 0.5 & 0.53 & 0.61 & 0.76 & 0.47 \\
U2 & 0.50 & 0.88 & 0.13 & 0.51 & 0.45 & 0.56 & 0.59 & 0.76 & 0.42 \\
U4 & 0.50 & 0.87 & 0.12 & 0.54 & 0.52 & 0.57 & 0.59 & 0.77 & 0.42 \\
SCAN & 0.51 & 0.95 & 0.08 & 0.57 & 0.53 & 0.62 & 0.82 & 0.90 & 0.74 \\
SCAN - \emph{Science} & 0.51 & 0.90 & 0.12 & 0.59 & 0.59 & 0.60 & 0.88 & 0.98 & 0.78 \\
SCAN - \emph{Metaphor} & 0.51 & 0.96 & 0.07 & 0.57 & 0.52 & 0.62 & 0.80 & 0.88 & 0.73 \\

\cline{1-10}

\end{tabular*}

\end{centering}

\end{table*}

\FloatBarrier

\begin{table*}[h]
\setlength\tabcolsep{3pt}
 \begin{centering}

\caption{BERT-cased: Accuracy on the Analogy Ranking Task}
\label{tab:table22}

\begin{tabular*}{\textwidth}{c||ccc}
\cline{1-4} 
\\
&BERT a-b non-tuned& Simple Classifier &
BERT a-b \\

\cline{1-4} 
OVERALL & 0.63 & 0.50 & 0.81 \\
\cline{1-4} 

SAT & 0.60 & 0.51 & 0.81 \\
U2 & 0.62 & 0.48 & 0.83 \\
U4 & 0.63 & 0.53 & 0.80 \\
SCAN & 0.63 & 0.49 & 0.81 \\
SCAN - \emph{Science} & 0.66 & 0.49 & 0.84 \\
SCAN - \emph{Metaphor} & 0.63 & 0.49 & 0.81 \\
\cline{1-4} 

\end{tabular*}
  \end{centering}
\end{table*}

\FloatBarrier

\begin{table*}[h]
\setlength\tabcolsep{3pt}
 \begin{centering}

\caption{BERT-large: Accuracy on the Analogy Ranking Task}
\label{tab:table23}

\begin{tabular*}{\textwidth}{c||ccc}
\cline{1-4} 
\\
&BERT a-b non-tuned& Simple Classifier &
BERT a-b \\

\cline{1-4} 
OVERALL & 0.66 & 0.53 & 0.68 \\
\cline{1-4} 

SAT & 0.65 & 0.55 & 0.64 \\
U2 & 0.67 & 0.53 & 0.67 \\
U4 & 0.67 & 0.52 & 0.67 \\
SCAN & 0.65 & 0.53 & 0.69 \\
SCAN - \emph{Science} & 0.64 & 0.54 & 0.72 \\
SCAN - \emph{Metaphor} & 0.65 & 0.53 & 0.68 \\
\cline{1-4} 

\end{tabular*}
  \end{centering}
\end{table*}

\FloatBarrier

\begin{table*}[h]
\setlength\tabcolsep{3pt}
 \begin{centering}

\caption{RoBERTa: Accuracy on the Analogy Ranking Task}
\label{tab:table24}

\begin{tabular*}{\textwidth}{c||ccc}
\cline{1-4} 
\\
&RoBERTa a-b non-tuned& Simple Classifier &
RoBERTa a-b \\

\cline{1-4} 

OVERALL & 0.60 & 0.52 & 0.88 \\
\cline{1-4} 

SAT & 0.59 & 0.52 & 0.86 \\
U2 & 0.65 & 0.56 & 0.86 \\
U4 & 0.61 & 0.53 & 0.89 \\
SCAN & 0.60 & 0.51 & 0.88 \\
SCAN - \emph{Science} & 0.60 & 0.53 & 0.89 \\
SCAN - \emph{Metaphor} & 0.60 & 0.50 & 0.88 \\

\cline{1-4} 

\end{tabular*}
  \end{centering}
\end{table*}

\FloatBarrier

\begin{table*}[h]
\setlength\tabcolsep{3pt}
\begin{centering}
\caption{BERT-cased: Accuracy on Distractor Dataset}
\label{tab:table25}

\begin{tabular*}{\textwidth}{@{\extracolsep{\fill}}cc||ccc|ccc|ccc}
\cline{1-11} 

\multicolumn{2}{c}{} &  \multicolumn{3}{c}{BERT a-b non-tuned} & \multicolumn{3}{c}{Simple Classifier}& \multicolumn{3}{c}{BERT a-b} \\
\cline{3-11} 
\multicolumn{1}{c}{Semantic } &
\multicolumn{1}{c}{Relation}& \multicolumn{3}{c}{Distractor Salience} & \multicolumn{3}{c}{Distractor Salience} & \multicolumn{3}{c}{Distractor Salience} \\

 Distance&Type& Overall & High &Low& Overall &  High &Low& Overall &  High &Low\\
\cline{1-11} 
 & Overall & 0.58 & 0.52 & 0.63 & 0.48 & 0.51 & 0.45 & 0.69 & 0.70 & 0.68 \\
 \cline{1-11} 

Near & Overall & 0.62 & 0.57 & 0.67 & 0.47 & 0.52 & 0.41 & 0.73 & 0.75 & 0.71 \\
 & Categorical & 0.70 & 0.70 & 0.70 & 0.61 & 0.77 & 0.44 & 0.70 & 0.65 & 0.75 \\
 & Causal & 0.40 & 0.30 & 0.50 & 0.52 & 0.55 & 0.48 & 0.66 & 0.68 & 0.64 \\
 & Compositional & 0.75 & 0.70 & 0.80 & 0.28 & 0.25 & 0.30 & 0.82 & 0.91 & 0.73 \\
Far & Overall & 0.53 & 0.47 & 0.60 & 0.50 & 0.50 & 0.49 & 0.65 & 0.65 & 0.64 \\
 & Categorical & 0.60 & 0.60 & 0.60 & 0.48 & 0.52 & 0.43 & 0.68 & 0.63 & 0.72 \\
 & Causal & 0.45 & 0.30 & 0.60 & 0.59 & 0.59 & 0.58 & 0.56 & 0.64 & 0.48 \\
 & Compositional & 0.55 & 0.50 & 0.60 & 0.44 & 0.40 & 0.47 & 0.70 & 0.67 & 0.73 \\
 \cline{1-11} 
\end{tabular*}
\end{centering}
\end{table*}

\begin{table*}[h]
\setlength\tabcolsep{3pt}
\begin{centering}
\caption{BERT-large: Accuracy on Distractor Dataset}
\label{tab:tables26}

\begin{tabular*}{\textwidth}{@{\extracolsep{\fill}}cc||ccc|ccc|ccc}
\cline{1-11} 

\multicolumn{2}{c}{} &  \multicolumn{3}{c}{BERT a-b non-tuned} & \multicolumn{3}{c}{Simple Classifier}& \multicolumn{3}{c}{BERT a-b} \\
\cline{3-11} 
\multicolumn{1}{c}{Semantic } &
\multicolumn{1}{c}{Relation}& \multicolumn{3}{c}{Distractor Salience} & \multicolumn{3}{c}{Distractor Salience} & \multicolumn{3}{c}{Distractor Salience} \\

 Distance&Type& Overall & High &Low& Overall &  High &Low& Overall &  High &Low\\
\cline{1-11} 
 & Overall & 0.58 & 0.55 & 0.60 & 0.57 & 0.60 & 0.54 & 0.57 & 0.58 & 0.56 \\
 \cline{1-11} 

Near & Overall & 0.63 & 0.57 & 0.70 & 0.59 & 0.59 & 0.59 & 0.61 & 0.60 & 0.61 \\

 & Categorical & 0.80 & 0.70 & 0.90 & 0.67 & 0.69 & 0.64 & 0.59 & 0.60 & 0.57 \\
 & Causal & 0.45 & 0.40 & 0.50 & 0.61 & 0.66 & 0.55 & 0.58 & 0.57 & 0.58 \\
 & Compositional & 0.65 & 0.60 & 0.70 & 0.50 & 0.42 & 0.57 & 0.66 & 0.63 & 0.68 \\
Far & Overall & 0.52 & 0.53 & 0.50 & 0.55 & 0.60 & 0.49 & 0.53 & 0.55 & 0.51 \\
 & Categorical & 0.65 & 0.50 & 0.80 & 0.58 & 0.6 & 0.56 & 0.60 & 0.62 & 0.58 \\
 & Causal & 0.50 & 0.50 & 0.50 & 0.53 & 0.64 & 0.41 & 0.57 & 0.58 & 0.55 \\
 & Compositional & 0.40 & 0.60 & 0.20 & 0.54 & 0.57 & 0.50 & 0.42 & 0.45 & 0.39 \\
 
 \cline{1-11} 

\end{tabular*}
\end{centering}
\end{table*}

\FloatBarrier

\begin{table*}
\setlength\tabcolsep{3pt}
\begin{centering}
\caption{RoBERTa: Accuracy on Distractor Dataset}
\label{tab:table27}

\begin{tabular*}{\textwidth}{@{\extracolsep{\fill}}cc||ccc|ccc|ccc}
\cline{1-11} 
\multicolumn{2}{c}{} &  \multicolumn{3}{c}{RoBERTa a-b non-tuned} & \multicolumn{3}{c}{Simple Classifier}& \multicolumn{3}{c}{RoBERTa a-b} \\
\cline{3-11} 

\multicolumn{1}{c}{Semantic } &
\multicolumn{1}{c}{Relation}& \multicolumn{3}{c}{Distractor Salience} & \multicolumn{3}{c}{Distractor Salience} & \multicolumn{3}{c}{Distractor Salience} \\
 Distance&Type& Overall & High &Low& Overall &  High &Low& Overall &  High &Low\\
\cline{1-11} 
 & Overall & 0.58 & 0.58 & 0.57 & 0.53 & 0.52 & 0.54 & 0.64 & 0.63 & 0.65 \\
 \cline{1-11} 

Near & Overall & 0.55 & 0.53 & 0.57 & 0.58 & 0.57 & 0.58 & 0.67 & 0.63 & 0.70 \\
 & Categorical & 0.55 & 0.60 & 0.50 & 0.60 & 0.61 & 0.59 & 0.60 & 0.60 & 0.60 \\
 & Causal & 0.65 & 0.60 & 0.70 & 0.55 & 0.56 & 0.53 & 0.80 & 0.70 & 0.90 \\
 & Compositional & 0.45 & 0.40 & 0.50 & 0.59 & 0.55 & 0.63 & 0.60 & 0.60 & 0.60 \\
Far & Overall & 0.60 & 0.63 & 0.57 & 0.49 & 0.47 & 0.50 & 0.62 & 0.63 & 0.60 \\
 & Categorical & 0.45 & 0.60 & 0.30 & 0.46 & 0.49 & 0.42 & 0.70 & 0.60 & 0.80 \\
 & Causal & 0.75 & 0.70 & 0.80 & 0.56 & 0.56 & 0.56 & 0.50 & 0.60 & 0.40 \\
 & Compositional & 0.60 & 0.60 & 0.60 & 0.44 & 0.35 & 0.53 & 0.65 & 0.70 & 0.60 \\
 
 \cline{1-11} 

\end{tabular*}
\end{centering}
\end{table*}

\end{document}